\documentclass[conference]{IEEEtran}
\IEEEoverridecommandlockouts
\usepackage{cite}
\usepackage{amsmath,amssymb,amsfonts}
\usepackage{algorithmic}
\usepackage{graphicx}
\usepackage{textcomp}
\usepackage{xcolor}
\def\BibTeX{{\rm B\kern-.05em{\sc i\kern-.025em b}\kern-.08em
    T\kern-.1667em\lower.7ex\hbox{E}\kern-.125emX}}

\usepackage[utf8]{inputenc}
\usepackage{mathtools}
\usepackage{amsthm, color}
\usepackage{bm, makecell}
\usepackage{tikz, subcaption, cite}
\usepackage{multirow}
\usepackage{array,booktabs,arydshln,xcolor}
\graphicspath{{fig/}}

\begin{document}

\title{A Deep-Learning-Aided Pipeline for Efficient Post-Silicon Tuning
\thanks{This research was supported by Advantest as part of the Graduate School ``Intelligent Methods for Test and Reliability'' (GS-IMTR) at the University of Stuttgart.}
}

\author{\IEEEauthorblockN{Yiwen Liao, Bin Yang}
\IEEEauthorblockA{\textit{Institute of Signal Processing and System Theory} \\
\textit{University of Stuttgart, Germany}\\
\{yiwen.liao, bin.yang\}@iss.uni-stuttgart.de}
\and
\IEEEauthorblockN{Jochen Rivoir, Rapha\"el Latty}
\IEEEauthorblockA{\textit{Applied Research and Venture Team} \\
\textit{Advantest Europe GmbH, Germany}\\
\{jochen.rivoir, raphael.latty\}@advantest.com}
}

\maketitle

\begin{abstract}
	In post-silicon validation, tuning is to find the values for the tuning knobs, potentially as a function of process parameters and/or known operating conditions. In this sense, an more efficient tuning requires identifying the most critical tuning knobs and process parameters in terms of a given figure-of-merit for a Device Under Test (DUT). This is often manually conducted by experienced experts. However, with increasingly complex chips, manual inspection on a large amount of raw variables has become more challenging. In this work, we leverage neural networks to efficiently select the most relevant variables and present a corresponding deep-learning-aided pipeline for efficient tuning.
\end{abstract}


\section{Introduction}
\label{sec:introduction}
The semiconductor industry has dramatically developed over the last few decades and forms the technology base for various applications in all aspects of daily life, from smart phones and personal computers to autonomous driving and Internet of Things (IoT)~\cite{mishra2019post,7892969}. To guarantee reliable performance and correct functionality of chip-based devices, validation plays a critical role. In particular, post-silicon validation (PSV) is known as one of the most complex and expensive components of the entire validation procedure for chip design because an actual fabricated device or chip is under test~\cite{mishra2019post}. 


Nowadays in PSV, fabricated semiconductor devices are often equipped with many ``tuning knobs'' in order to counteract the effect of process variations and mitigate non-ideal designs. Moreover, the number of tuning knobs is notably growing due to the finer structures and complex manufacturing process. This leads to high cost in time and difficulty in analysis for experts. Accordingly, there is a pressing need for an intelligent method to help experts identify the most representative tuning knobs and process parameters for a given figure-of-merit (FoM). As a result, it is natural to introduce variable selection~\cite{guyon2003introduction} to identify the most critical and relevant tuning knobs and process parameters from hundreds of candidates. However, conventional approaches cannot scale to a large number of candidate variables, or fail to model the nonlinear relations between the candidate variables and FoM. Last but not least, mixed data types including categorical and numerical values can be another challenge for PSV as a whole, and for conventional (variable selection) approaches in particular.

\begin{figure}[!ht]
	\centering
	\begin{tikzpicture}[scale=0.75]
	\draw[thick, black!55, rounded corners=1mm,fill=blue!15, fill opacity=0.5] (-1.5, 0.5) -- (-1.5, -1.5) -- (1.5, -1.5) -- (1.5, 0.5) -- cycle;
	
	\draw[thick, black!55, rounded corners=1mm,fill=orange!15, fill opacity=0.5] (2.25, 0.5) -- (5.75, 0.5) -- (5.75, -1.5) -- (2.25, -1.5) -- cycle;
	
	\draw[thick, black!55, rounded corners=1mm,fill=green!15, fill opacity=0.5] (6.5, 0.5) -- (9., 0.5) -- (9., -1.5) -- (6.5, -1.5) -- cycle;
	
	\node[rectangle, draw, rounded corners=1mm, rotate=0] (dut) at(0, 0) {DUT};
	\node[rectangle, draw, rounded corners=1mm, rotate=0] (res) at(0, -1) {test cases};
	
	\node[rectangle, draw, rounded corners=1mm, rotate=0] (ml) at(4, 0) {DL algorithm};
	\node[rectangle, draw, rounded corners=1mm, rotate=0] (fs) at(4, -1) {selection};
	
	\node[rectangle, draw, rounded corners=1mm] (downstream) at(7.75, -0.5) {analysis};
	
	\draw[thick, black!55, ->] (dut)--(res);
	\draw[thick, black!55, ->] (res) to[out=0,in=180] (ml);
	\draw[thick, black!55, ->] (ml)--(fs);
	\draw[thick, black!55, ->] (fs) to[out=0,in=180] (downstream);
\end{tikzpicture}
	\caption{The DL-aided pipeline for post-silicon tuning.}
	\label{fig:pipeline}
\end{figure}
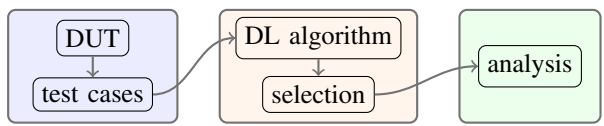

To address the challenges above, in this work, we leverage deep learning (DL) techniques to select the most crucial variables during PSV. Our experiments show that the new pipeline can scale to hundreds of candidate variables and is able to identify hidden interactions between variables.

\section{Methodology}
\label{sec:method}

The key idea of the proposed DL-aided pipeline for tuning in post-silicon validation is to introduce a variable selection approach based on neural networks (NN). As shown in Fig.~\ref{fig:pipeline}, the new pipeline is conspicuous due to the deep learning block (\emph{orange}) between the testing devices (\emph{blue}) and analysis by experts (\emph{green}). Specifically, the DL-block is trained to identify the most informative and representative candidate variables from the test cases in an efficient way and forward the selected variables only to the analysis block. Within the analysis block, the experts cares about the selected variables only and do not need to deal with high-dimensional data anymore, which enables more efficient and reliable visualization and modeling in PSV, e.g. for a contemporary work in this workshop~\cite{domanski2021self}.

In the proposed pipeline, we use our recently published FM-method~\cite{9533531} as the backbone algorithm. The novel FM-method is an end-to-end variable selection approach and can be easily integrated to the DL-based pipeline as follows.

\subsubsection{Test Cases Preprocessing}
\begin{figure}[!ht]
	\centering
	\begin{tikzpicture}[scale=0.425]
	\definecolor{mycolor1}{RGB}{215,48,39}
	\definecolor{mycolor2}{RGB}{255,255,191}
	\definecolor{mycolor3}{RGB}{123,50,148}
	
	\draw[thick, draw=black!55, dashed, rounded corners=1mm,fill=orange!15,fill opacity=0.3] (-6, 3.) -- (-6, -3.5) -- (10, -3.5) -- (10, 3.) -- cycle;
	
	\node[rotate=90] (test) at (-7, 0) {test cases};
	\node[rotate=90] (processing) at (-5, 0) {preprocessing};
	\node (x) at(-3, 0) {$X$};
	\node (m) at(3, -2.5) {$\bm{m}$};
	\node (odot) at (3, 0) {$\odot$};
	\node (y) at(9.5, 0) {$\hat{Y}$};
	\node (analysis) at(6, -4.75) {analysis};
	
	\draw[thick, draw=black!55, rounded corners=0.5mm,fill=mycolor1,fill opacity=0.3] (4, -1.5) -- (4, 1.5) -- (8, 0.8) -- (8, -0.8) -- cycle;
	\draw[thick, draw=black!55, rounded corners=0.5mm,fill=mycolor3,fill opacity=0.3] (1.5, -1.75) -- (1.5, -3.25) -- (-1, -3.25) -- (-1, -1.75) -- cycle;
	
	\draw[thick, black!55, ->] (test) -- (processing);
	\draw[thick, black!55, ->] (processing) -- (x);
	\draw[thick, black!55, ->] (x) -- (-2, 0) -- (-2, -2.5) -- (-1, -2.5);
	\draw[thick, black!55, ->] (1.5, -2.5) -- (m);
	\draw[thick, black!55, ->] (m) -- (odot);
	\draw[thick, dotted, black!55, ->] (m) -- (6, -2.5) -- (analysis);
	\draw[thick, black!55, ->] (x) -- (odot);
	\draw[thick, black!55, ->] (odot) -- (4, 0);
	\draw[thick, black!55, ->] (8, 0) -- (y);
	\draw[fill=black!55, draw=black!55] (-2, 0) circle(0.1);
	
	\node (g) at(6, 0) {NN};
	\node (f) at(0.25, -2.5) {FM};

%
%
%
%
%
%
\end{tikzpicture}
	\caption{The structure of the FM-method for variable selection.}
	\label{fig:fm}
\end{figure}
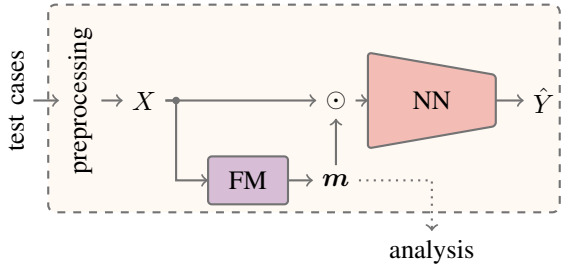

In PSV, the test cases are often stored in tabular format which cannot be directly processed by neural networks and require preprocessing. Therefore, as shown in Fig.~\ref{fig:fm}, a preprocessing step is added before the FM-module so that the raw categorical candidate variables are firstly converted to numerical forms (e.g. each category is encoded into an integer number). Subsequently, all candidate variables are normalized to a similar scale for an efficient training in the next steps and the resulting data are denoted as $X$.

\subsubsection{Joint Training of the FM-Module and Neural Network}
During training, $X$ is fed into the FM-module in small batches and a corresponding unique feature mask $\bm{m}$ is generated during each iteration, where the dimension of $\bm{m}$ is the same as the total number of candidate variables. At each iteration, the generated $\bm{m}$ is element-wisely multiplied to the input data batch and the product is then fed into a neural network (NN) in order to output the prediction $\hat{Y}$. Accordingly, by minimizing a loss between the prediction $\hat{Y}$ and the ground truth $Y$ (i.e. the target variables), both FM-module and NN are jointly trained.

\subsubsection{Feature Mask Generation and Selection}
After training, the entire training data $X$ is fed to the trained FM-module and one unique $\bm{m}$ can be therefore obtained, where each element of $\bm{m}$ indicates the importance of the corresponding candidate variable. The most critical candidate variables can be selected based on the learned importance scores.


It should be noted that the proposed pipeline is generic; e.g. the preprocessing approach and the loss function can be defined according to different use cases or requirements by users and practitioners.

\section{Experiments}
\label{sec:exp}
This section justifies the relevance of applying DL algorithms to post-silicon tuning. Specifically, we used a real-world dataset from Advantest consisting of 100,000 test cases from a single DUT. The entire dataset was converted to a matrix, where each row denotes a test case (training sample) with 11 dimensions (candidate variables) and a continuous FoM value as its label (target variable). Specifically, the candidate variables consist of 7 tuning knobs (t1 to t7) and 4 operation conditions (c1 to c4). Accordingly, in the use case of tuning, semiconductor experts attempt to identify the most important and critical candidate variables that can as accurately as possible predict the target variable. Note that this dataset consists of mixed data types (i.e. categorical and numerical candidate variables). Fortunately, neural networks can deal with different data types with simple encoding. In our experiments, we encoded all categorical variables into positive integers (i.e. $1, 2, 3, \dots$) and subsequently normalized them into the range $[0, 1]$ using minmax-scaling.

To apply our pipeline to this dataset, we implemented the variable selection algorithm as follows. The learning network consisted of two hidden dense layers with 64 and 32 neurons respectively and both layers used a LeakyReLU~\cite{maas2013rectifier} as activation with a rate of 0.02. The output layer had one neuron with a linear activation since the variable selection was performed in combination with a regression task, i.e. the prediction of a continuous numeric target variable. As a result, the minimization of the mean squared error loss was used as the overall learning objective.

\begin{figure}[!ht]
	\centering
	\includegraphics[width=\linewidth]{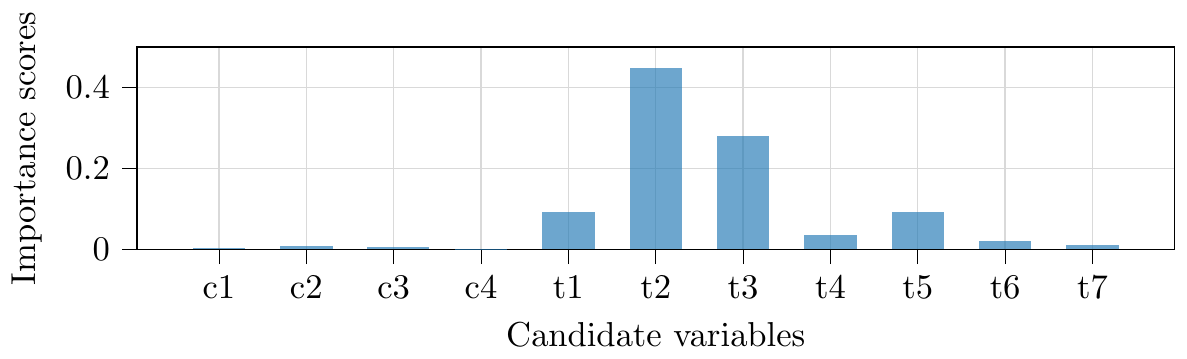}
	\caption{The learned variable importance scores for the candidate variables.}
	\label{fig:res}
\end{figure}

After training, the FM-module generates the learned importance scores (i.e. $\bm{m}$) for all 11 candidate variables, where greater scores indicate more importance for predicting the target variable. Fig.~\ref{fig:res} shows the learned feature importance scores for the 11 candidate variables. It is clear to see that t1 to t5 have the greatest importance scores and are therefore considered as the five most important variables for tuning. Meanwhile, none of the 4 operation conditions were relevant for this DUT. The selection result matches the exhaustive search and shows the effectiveness of the proposed pipeline.

\begin{figure}[!ht]
	\centering
	\includegraphics[width=\linewidth]{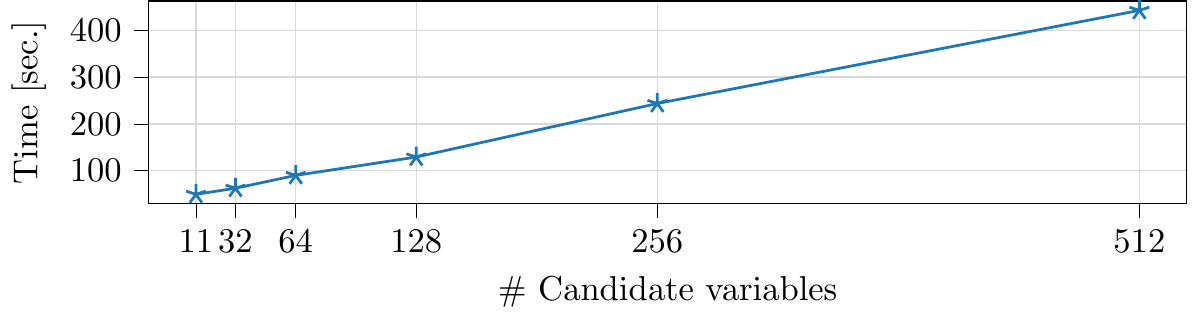}
	\caption{Time consumption vs. number of candidate variables.}
	\label{fig:time_vs_var}
\end{figure}

Furthermore, thanks to the parallel computation on one graphics processing unit (GPU), the total training time was less than one minute for the entire dataset with 100000 samples. As comparison, an exhaustive search based on conventional statistical approaches can take hours. Specifically, Fig.~\ref{fig:time_vs_var} shows the time consumption over different numbers of candidate variables\footnote{We added independent artificial variables to the original dataset to obtain different numbers of candidate variables in order to justify the time consumption of our method.}. It can be observed that our method has only an almost linear complexity with respect to the number of candidate variables, while conventional statistical exhaustive search methods typically have exponential complexity. The efficiency of our method reveals the significant potential of integrating our novel pipeline to many existing analysis tools for post-silicon validation.

Moreover, in contrast to conventional statistical variable selection approaches, our method can be easily extended according to specific requirements. One typical use case is that the figure-of-merit can be multivariate. Conventional methods may require different models for different target variables and aggregate the final results based on expert knowledge. On the contrary, this procedure can be significantly simplified by using neural networks. Broadly speaking, neural networks allow multivariate outputs or multiple output layers. This can enable an end-to-end variable selection with respect to multivariate regression, which is expected to accelerate the analysis process in practice.


\section{Discussion}
\label{sec:discussion}

Based on the conducted experiments, it is clear that applying deep learning to post-silicon validation is of great interest and shows promising results. However, we have observed a challenging corner case for DL-based pipeline in post-silicon validation. More precisely, our approach as well as other DL-based variables selection methods generate one unique importance score vector after training and then selects the largest elements. This means that a selected variable subset with the size of $L_1$ consists of all elements of a smaller variable subset with the size of $L_2$, where $L_2 < L_1$. On one hand, this allows an efficient selection for many situations; i.e. training once can provide compact selection results. On the other hand, this mechanism is not able to provide correct selection when one candidate variable is only necessary in combination with another variable. To address this corner case is left as a future work by us.


\section{Conclusion}
\label{sec:conclusion}
This paper proposes a deep-learning-based pipeline for tuning in post-silicon validation. By introducing deep learning to variable selection, domain experts can avoid dealing with high-dimensional data, which leads to a more efficient and reliable analysis on the test cases of DUTs. Furthermore, DL-based approaches allow easy extensions w.r.t. specific requirements in different use cases. As future work, we plan to automate the hyperparameter optimization procedure within the DL-block to make the pipeline more friendly to users who are not familiar with deep learning.

\bibliographystyle{IEEEbib}
\bibliography{Refs}

\end{document}